\PassOptionsToPackage{table,xcdraw,dvipsnames}{xcolor}
\documentclass[sigconf,natbib=true]{acmart}
\AtBeginDocument{%
  }

\setcopyright{acmlicensed}
\copyrightyear{2018}
\acmYear{2018}
\acmDOI{XXXXXXX.XXXXXXX}
\acmConference[Conference acronym 'XX]{Make sure to enter the correct
  conference title from your rights confirmation email}{June 03--05,
  2018}{Woodstock, NY}
\acmISBN{978-1-4503-XXXX-X/2018/06}

\makeatletter
\newif\if@restonecol
\makeatother

\usepackage{algorithm}
\usepackage{multirow}
\usepackage{algpseudocode}
\usepackage{colortbl}
\usepackage{tabularx}
\usepackage{booktabs}
\usepackage[normalem]{ulem}

\algrenewcommand\algorithmicrequire{\textbf{Input:}}
\algrenewcommand\algorithmicensure{\textbf{Output:}}

\begin{document}

\title{Enabling Collaborative Parametric Knowledge Calibration for Retrieval-Augmented Vision Question Answering}

\author{Jiaqi Deng}
\email{Jiaqi.Deng@student.uts.edu.au}
\orcid{0009-0008-5426-127X}
\affiliation{%
  \institution{University of Technology Sydney}
  \city{Sydney}
  \state{New South Wales}
  \country{Australia}
}

\author{Kaize Shi}
\email{Kaize.Shi@uts.edu.au}
\affiliation{%
  \institution{University of Technology Sydney}
  \city{Sydney}
  \state{New South Wales}
  \country{Australia}
}

\author{Zonghan Wu}
\email{zhwu@sem.ecnu.edu.cn}
\affiliation{%
  \institution{East China Normal University}
  \city{Shanghai}
  \country{China}
}

\author{Huan Huo}
\email{Huan.Huo@uts.edu.au}
\affiliation{%
  \institution{University of Technology Sydney}
  \city{Sydney}
  \state{New South Wales}
  \country{Australia}
}

\author{Dingxian Wang}
\email{Dingxian.Wang@student.uts.edu.au}
\affiliation{%
  \institution{University of Technology Sydney}
  \city{Sydney}
  \state{New South Wales}
  \country{Australia}
}

\author{Guandong Xu}
\email{gdxu@eduhk.hk}
\affiliation{%
  \institution{The Education University of Hong Kong}
  \city{Hong Kong}
  \state{Hong Kong}
  \country{China}
}

\renewcommand{\shortauthors}{Deng et al.}

\begin{abstract}
    Knowledge-based Vision Question Answering (KB-VQA) systems address complex visual-grounded questions with knowledge retrieved from external knowledge bases. The tasks of knowledge retrieval and answer generation tasks both necessitate precise multimodal understanding of question context and external knowledge. However, existing methods treat these two stages as separate modules with limited interaction during training, which hinders bi-directional parametric knowledge sharing, ultimately leading to suboptimal performance. To fully exploit the cross-task synergy in KB-VQA, we propose a unified retrieval-augmented VQA framework with collaborative parametric knowledge calibration. The proposed framework can effectively adapt general multimodal pre-trained models for fine-grained, knowledge-intensive tasks while enabling the retriever and generator to collaboratively enhance and share their parametric knowledge during both training and inference. To enhance fine-grained understanding of questions and external documents, we also integrate late interaction mechanism into the proposed training framework. Additionally, we introduce a reflective-answering mechanism that allows the model to explicitly evaluate and refine its knowledge boundary. Our approach achieves competitive performance against state-of-the-art models, delivering a significant 4.7\% improvement in answering accuracy, and brings an average 7.5\% boost in base MLLMs' VQA performance. The code is available at https://anonymous.4open.science/r/UniRVQA-D8C7.
\end{abstract}



\received{20 February 2007}
\received[revised]{12 March 2009}
\received[accepted]{5 June 2009}

\maketitle

\section{Introduction}
Knowledge-based Vision Question Answering (KB-VQA) is a task of answering challenging image-grounded questions that require the integration of external knowledge beyond commonsense, such as web-sourced encyclopedic documents~\cite{Vrandecic2014Wikidata, Luo2021Weakly-SupervisedAnswering}. Recent advances in large multimodal models demonstrate strong capabilities in leveraging the vast implicit knowledge stored in their parameters~\cite{Yang2022AnVQA,Alayrac2022Flamingo:Learning, Chen2023PaLI:Model,Ma2024GeReA:Answering}. However, these large models may struggle with timely knowledge updates, factual errors and private applications~\cite{Mallen2023WhenMemories, Min2023FActScore:Generation}. Therefore, retrieval-augmented methods have emerged as an alternative solution to the KB-VQA task, which support knowledge-intensive generation with the knowledge retrieved from external sources, providing the generator with relevant context and reducing the need for large-scale tuning.

\begin{figure}
    \centering  \includegraphics[width=\columnwidth]{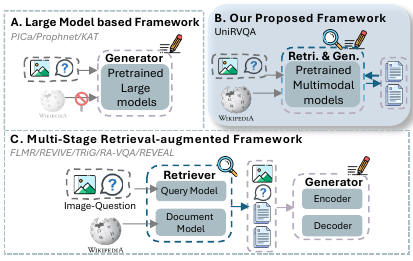}
    \caption{A comparison between the large-model-based framework, multi-stage retrieval-augmented framework, and our proposed unified framework (UniRVQA). The representative systems are listed in gray texts for exemplifications.}
    \label{figure1}
\end{figure}

Existing work on retrieval-augmented KB-VQA often employs sequential and independent models for the knowledge retriever and answer generator~\cite{Gui2021KAT:Vision-and-Language,Lin2022REVIVE:Answering, Lin2023Fine-grainedAnswering}. Such modular separation inherently prevents the retriever and generator from benefiting each other's training process and sharing parametric knowledge. Recent efforts ~\cite{Lin2022RetrievalKnowledge,Siriwardhana2023ImprovingAnswering} have explored optimizing the retriever with guidance from the generator. However, these methods still operate under a unidirectional knowledge passage: while the retriever’s parametric knowledge is refined through the generator’s training signals, no reciprocal feedback is provided to the generator. On the other hand, knowledge retrieval and answer generation both require precise understanding and reasoning over question context and external knowledge documents. Therefore, we argue that enabling bidirectional parametric knowledge sharing during training and inference could better exploit the interdependence across tasks and improve system performance.

The other challenge of KB-VQA lies in the fine-grained multimodal representation learning. Earlier verbalization approaches reformulate VQA into textual QA by converting visual inputs into text via image captioning or dense labeling~\cite{Lin2022REVIVE:Answering, Gui2021KAT:Vision-and-Language, Gao2022Transform-Retrieve-Generate:Answering}. However, such transformation often discards complex visual details, thereby limiting the ability in context understanding and downstream knowledge retrieval~\cite{Ma2024GeReA:Answering, Lin2023Fine-grainedAnswering}. Recent advancements in multimodal alignment~\cite{Radford2021LearningSupervision, Li2022BLIP:Generation} have empowered general pre-trained Multimodal Large Language Models (MLLMs), such as BLIP2~\cite{Li2023BLIP-2:Models} and InstructBLIP~\cite{Dai2023InstructBLIP:Tuning}, which excel at answering general visual questions. However, these models can struggle with knowledge-intensive tasks due to limited capacity for capturing fine-grained semantic nuances. While specialized pre-trained models have been proposed for knowledge-based retrieval and VQA~\cite{Chen2022MuRAG:Text,Yasunaga2022Retrieval-AugmentedModeling,Hu2022REVEAL:Memory,Lin2023Fine-grainedAnswering, Lin2024PreFLMR:Retrievers}, it would be highly beneficial to adapt existing general pre-trained multimodal models for retrieval-augmented KB-VQA as they have already memorized extensive general knowledge. Such adaptation requires much fewer computational resources compared to pre-trained specialized models, yet it remains underexplored. 

To address the challenge of insufficient parametric knowledge sharing and ineffective fine-grained multimodal representation, this paper proposes a \textbf{Uni}fied \textbf{R}etrieval-Augmented \textbf{V}ision \textbf{Q}uestion \textbf{A}nswer (\textbf{UniRVQA}) framework. As demonstrated in Fig.\ref{figure1}-B, our framework is designed to calibrate the knowledge space of the retriever and the answer generator through a joint optimization scheme. The proposed setup fosters collaborative parametric knowledge learning, thereby overcoming modular isolation and enabling both components to benefit from each other’s learning process. Simultaneously, the integration of a late interaction mechanism~\cite{Khattab2020ColBERT:BERT} further enhances the system’s ability to capture fine-grained semantic alignments between modalities. With an additional novel reflective-answering mechanism, the model learns not only to generate answers using retrieved knowledge, but also to assess the correctness of its own responses and internally evaluate its knowledge boundaries. The main contributions of this paper are summarized as below:

\begin{itemize}
    \item We propose a novel KB-VQA framework with collaborative parametric knowledge calibration across tasks. The framework adapts general MLLMs to effectively handle both tasks along the KB-VQA pipeline, including fine-grained knowledge retrieval and visual question answering.
    \item We introduce a reflective-answering mechanism for model to assess its knowledge boundaries, alongside incorporating the late-interaction mechanism into the unified framework to enhance fine-grained multimodal representation.
    \item Experiments on two public benchmarks show that our model outperforms state-of-the-art methods with significant improvements of 4.78\% and improves base MLLMs by an average of 7.54\% in answering accuracy.
    
\end{itemize}

\section{Relate Work}

\subsection{Knowledge-based VQA Systems}
KB-VQA systems~\cite{Yang2022AnVQA, Lin2022RetrievalKnowledge, Lin2023Fine-grainedAnswering, Yu2023Prophet:Answering, Salemi2023AAnswering} solve complex knowledge-intensive visual questions. One of the approaches to KB-VQA is to leverage implicit knowledge from large language models, such as GPT-3~\cite{Brown2020LanguageLearners}, using carefully crafted prompts. For example, KAT~\cite{Gui2021KAT:Vision-and-Language} and PICa~\cite{Yang2022AnVQA} transform images into textual captions so that the visual context can be utilized as part of the prompts by GPT-3\cite{Brown2020LanguageLearners}. However, these large models are less adaptable when updating with new knowledge and often fail to mitigate their inherent factual errors~\cite{Mallen2023WhenMemories, Min2023FActScore:Generation}. Therefore, the retrieval-augmented generation (RAG) paradigm has emerged as a more efficient and flexible alternative for KB-VQA. RAG-based systems first retrieve relevant knowledge, which is then processed by the generator for answer generation. These external knowledge can either come from structured Knowledge Graphs~\cite{Li2020BoostingAggregation, Guo2022AVQA, Speer2016ConceptNetKnowledge} or unstructured documents such as documents on Wikipedia ~\cite{Vrandecic2014Wikidata,Lin2023Fine-grainedAnswering}.  

Existing works on retrieval-augmented KB-VQA often have independent retriever and generator models that are trained separately~\cite{Lin2022REVIVE:Answering, Lin2022RetrievalKnowledge,Lin2023Fine-grainedAnswering, Hu2022REVEAL:Memory}. For example, REVIVE~\cite{Lin2022REVIVE:Answering} adopts a pre-trained CLIP~\cite{Liu2023LearningKnowledge} to extract features for retrieval and adopt multiple FiD networks ~\cite{Izacard2021LeveragingAnswering} as the backbone of the generator. However, the independence of the generator and retriever can lead to compromised performance as their tasks are closely related. To mitigate the issue, RA-VQA~\cite{Lin2022RetrievalKnowledge} propose to first train the retrieval network, followed by refining the retriever under the guidance of generator's training signal. However, cross-task interactions and parametric knowledge sharing are still constraint with the separated modules. Therefore, the generator is blind to the retriever's evolvement. Given the interdependence across KB-VQA tasks, we suggest that it would be beneficial to have a more integrated framework as both tasks need a fine-grained understanding of knowledge and questions. Moreover, existing work~\cite{Lin2022RetrievalKnowledge,Yang2022AnVQA, Gao2022Transform-Retrieve-Generate:Answering} often verbalize images by applying image-to-text transformation~\cite{Shi2017AnRecognition, Anderson2018Bottom-UpAnswering, Li2020Oscar:Tasks}, which often results in the loss of critical fine-grained visual information.

\subsection{Pre-trained Multimodal Models in KB-VQA} 

Recent large vision-language pre-trained models ~\cite{Radford2021LearningSupervision, Dai2023InstructBLIP:Tuning, Li2023BLIP-2:Models} offer an effective solution to the above challenge brought by verbalization, with its superior capabilities in multimodal representation learning. These general MLLMs are usually pre-trained on large-scale datasets so that knowledge can be stored in model parameters. Therefore, MLLMs like BLIP2~\cite{Li2023BLIP-2:Models} have a superior visual understanding capability, by constructing a lightweight Querying Transformer between visual encoders (e.g. ViT-L/14~\cite{Dosovitskiy2020AnScale}) and LLMs (e.g. Flan-T5~\cite{Raffel2019ExploringTransformer}) 
in a wide range of downstream multimodal tasks, such as image-text retrieval and image-grounded question answering. However, they still struggle with document retrieval for KB-VQA and, therefore, fail to elevate themselves to a more satisfactory answering performance. This is because these general pre-trained models are not primarily trained to capture fine-grained nuances within external documents.

\begin{figure*}[htbp]
    \centering  \includegraphics[width=\textwidth]{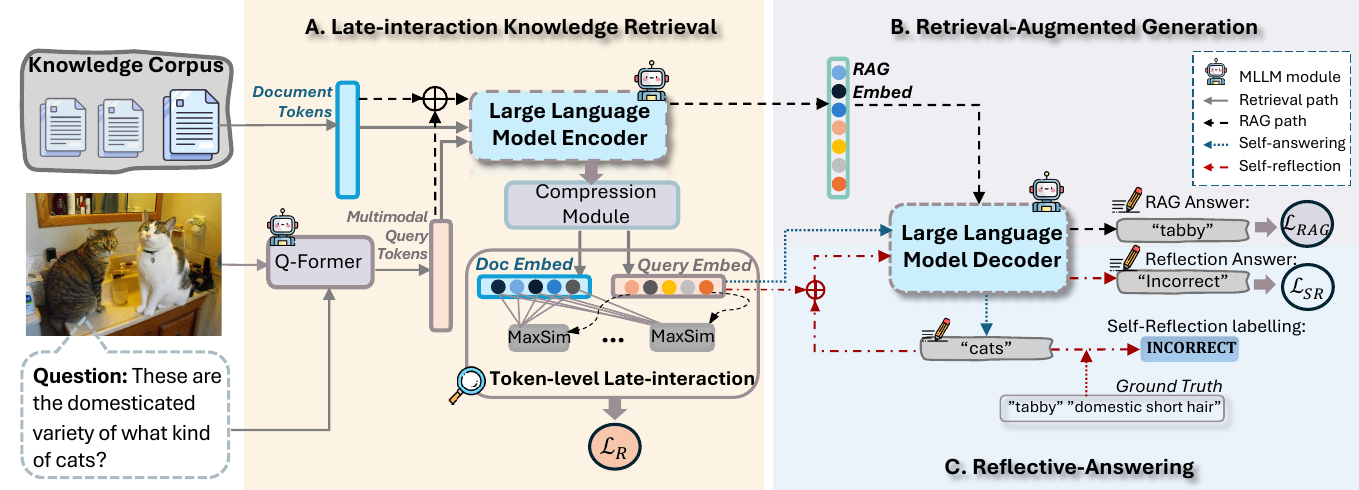}
    \caption{An overview of the proposed Unified Retrieval-Augmented Vision Question Answering  framework (UniRVQA). The framework consists of two main pathways: (1) Part A and B perform late-interaction retrieval and retrieval-augmented generation, which together form the RAG path. (2) Part C outlines the reflective-answering mechanism, where the base model conducts self-answering and evaluates the correctness simultaneously.}
    \label{fig:enter-label}
\end{figure*}

To address this issue, recent efforts have focused on specialized pre-trained models to enhance their knowledge-intensive retrieval and answering capabilities. For instance, FLMR~\cite{Lin2023Fine-grainedAnswering} designs a mapping layer to project visual embeddings to token-level language embeddings for downstream retrieval tasks. MuRAG~\cite{Chen2022MuRAG:Text} trains a cross-modal transformer to fuse the visual and textual embeddings. The model undergoes pre-training on a large-scale dataset that integrates images, text, and knowledge by applying the joint learning strategy. Similarly, RA-CM3~\cite{Yasunaga2022Retrieval-AugmentedModeling} also injects knowledge bases during the pre-training process to align image-text-knowledge tuples so that the model can be equipped with knowledge retrieval capabilities. Despite the relatively promising performance these models have achieved, we argue that these pre-training methods are computationally expensive and less flexible to update with latest information. Given the rich parameterized knowledge of general MLLMs, it would be worthwhile to adapt them for their potential in KB-VQA, which is a non-trivial yet sparsely researched question.

\section{Methodology}

\subsection{Problem Formulation} 
We consider the general setting of KB-VQA for the framework design. Given a textual question $Q$ regarding an image $I$, the objective of KB-VQA is to generate an answer $\hat{a}$ based on retrieved relevant documents $\mathcal{D} = \{d_i\}^k_{i=1}$:

\begin{equation}
\label{eqn1}
    \hat{a} = \arg\max_{a, d_i\in\mathcal{D}} p_{\Phi}(a |Q, I, d_i),\\
\end{equation}
where $\Phi$ denotes the parameters of the base model. The answer is based on the Bayesian joint probability of retrieval and generation:
\begin{equation}
\label{eqn2}
    p(a |Q, I, \mathcal{D}_{full})=\underbrace{p_{\phi}(\mathcal{D} |Q, I, \mathcal{D}_{full})}_\text{Retrieval}\cdot 
   \underbrace{p_{\Phi}(a |Q, I, \mathcal{D})}_\text{Generation},
\end{equation}
where $\mathcal{D}_{full}$ represents the external knowledge base of size $N$ and $\phi$ denotes the parameters for retrieval models. In the UniRVQA setting, the number of retrieved documents $k \ll N$ and $\phi  \subseteq \Phi$.

\subsection{Unified Multimodal Embedding} 

With the pre-trained MLLM as our base model, we can first obtain the unified embeddings from both textual and visual input to construct a query embedding $\mathcal{\tilde{Q}}$: 

\begin{equation}
    \mathcal{\tilde{Q}} =[f_{mm}(Q), f_{mm}(I)]\in \mathbb{R}^{l_{Q}\times h}\\
\end{equation}
where $h$ is the hidden size and $l_{Q}$ is the total length of the sequence by concatenating image tokens and text tokens embeddings. $f_{mm}$ is a part of the pre-trained MLLM that generates semantically-meaningful embeddings. In our framework, we adopt BLIP2~\cite{Li2023BLIP-2:Models} or InstructBLIP~\cite{Dai2023InstructBLIP:Tuning} as the base model, where $f_{mm}$ consists of a pre-trained Q-Former and a T5 encoder. With the same encoder, we can obtain the embedding of the document $d$ with a length $l_{d}$ in the external knowledge base:

\begin{equation}
    \mathcal{\tilde{D}} =[f_{mm}(d))]\in \mathbb{R}^{l_{d}\times h}\\
\end{equation}
\subsection{Collaborative Parametric Knowledge Calibration Training Framework}
To fully utilize the interplay across retrieval and generation tasks in KB-VQA, we propose a unified framework, where the retriever and the answer generator share the parametric knowledge and mutually calibrate each other during training. Specifically, the training framework consists the simultaneous learning processes of three main tasks: late-interaction knowledge retrieval, retrieval-augmented generation and reflective answering.
\subsubsection{Late-Interaction Knowledge Retrieval} 
During the retrieval stage, the relevance $r$ between a particular query $\mathcal{\tilde{Q}}$ and a document $\mathcal{\tilde{D}}$ is assessed using a relevance score in a late-interaction manner, following ColBERT~\cite{Khattab2020ColBERT:BERT}. Late-interaction retrieval is known as a fine-grained and more efficient approach, where query and document representations are independently encoded before interacting~\cite{Khattab2020ColBERT:BERT}. We extend ColBERT from a individual retrieval model to a flexible module compatible with any MLLM-based framework, allowing end-to-end training with the generation task.

\begin{equation}
    r(\mathcal{\tilde{Q}}, \mathcal{\tilde{D}}) =\sum_{i=1}^{l_q}\max_{j=1}^{l_d}\mathcal{\tilde{Q}}_i\mathcal{\tilde{D}}_j^T
\end{equation}

The relevance score is calculated based on the token-level embeddings. For each token in the query, the document token with the highest relevance score will be identified, and these maximum scores will then be summed up to produce the overall relevance score. Our empirical study highlights that an additional step is needed to extend the use of ColBERT~\cite{Khattab2020ColBERT:BERT}, as the hidden size of the output from the MLLMs encoder is relatively large to conduct relevance calculation and indexing. To improve the efficiency of the retriever, we adopt a simple yet effective compression module that packs token embeddings into lower-dimensional latent spaces by using two multi-layer perceptron layers, connected by a ReLU activation function. 
 
To elicit the retrieval ability of pre-trained MLLMs by learning the fine-grained relevance of the query and documents, we adopt an in-batch contrastive learning strategy, following~\cite{Karpukhin2020DenseAnswering, Luo2021Weakly-SupervisedAnswering, Lin2023Fine-grainedAnswering}. Given each query $\mathcal{\tilde{Q}}$ in the batch, the ground-truth positive documents for other queries in the same batch will be considered as its negative samples, denoted as $D_n$. The contrastive learning retrieval loss will be formulated as:
\begin{equation} 
\mathcal{L}_{R}=-\sum_{\mathcal{\tilde{Q}}, \mathcal{\tilde{D}}^+} \log \frac{\exp(r(\mathcal{\tilde{Q}}, \mathcal{\tilde{D^+}}))}{\exp(r(\mathcal{\tilde{Q}}, \mathcal{\tilde{D}}^+)) + \sum_{\mathcal{\tilde{D}}^+\in D_n} \exp(r(\mathcal{\tilde{Q}}, \mathcal{\tilde{D}}^+))}
\end{equation}

\subsubsection{Retrieval-Augmented Generation}

After the late-interaction relevance calculation, the embeddings of the positive document will be concatenated with the query embeddings and fed to the language model for answer generation. The answer generation will be trained by the casual language modeling loss:
\begin{equation}
\label{eqn8}
\mathcal{L}_{RAG}=-\sum_{\mathcal{\tilde{Q}}, \mathcal{\tilde{D}}^+}\log p_\Phi(a|\mathcal{\tilde{Q}}, \mathcal{\tilde{D^+}}),
\end{equation}
 For each open-ended question, there will be a set of human responses $\mathcal{A}$. The target answer $a$ will be randomly selected from the set. Our empirical studies show that random selection enhances the model's robustness to document noise compared to using the answer provided in the document. Based on Eq.~\ref{eqn2}, the two loss terms can be directly summed together as the log-joint probability of retrieval and generation:

\begin{equation}
\label{equ8}
\mathcal{L}_{RAG\_joint}=\mathcal{L}_{R}+\mathcal{L}_{RAG}
\end{equation}

\begin{algorithm}
\caption{Pseudo Code of UniRVQA Reflective Answering Training Process}\label{alg:sr}
\begin{algorithmic}[1]
\Require MLLM (Encoder $\mathcal{E}$, Decoder$\mathcal{D}$, Parameters $\Phi$); Question-image pairs $q$; batch $\mathcal{B}$ with size $n$; Target Answer Set $\mathcal{A}$; Learning rate $\alpha$
\Ensure{$\mathcal{L}_{SR}=\mathcal{L}_{gen}+\mathcal{L}_{reflect}$}

\For{t=1,2,...,T}
\For{each image-question pair ($q_i$) in $\mathcal{B}_t^n$}
\State Predict answer: $\hat{a_i} = \mathcal{D}(\mathcal{E}(q_i))$
\State Calculate the loss for the self-generation:
\State $\mathcal{L}_{SR}=\mathcal{L}_{gen}= \sum_{i}^n\textsc{CrossEntropyLoss}(\hat{a_i}, \mathcal{A}_i)$
\While {$t \geq s_{join}$}\Comment{Late join}
\State Construct self-reflection label $r_i$:
\If {$\hat{a_i} \in \mathcal{A}_i$}: $r_i \mathrel{==}\textcolor{MidnightBlue}{\textsc{"correct"}}$
\Else{: $r_i \mathrel{==}\textcolor{BrickRed}{\textsc{"incorrect"}}$}
\EndIf
\State Predict self-reflection result: $\hat{r_i} = \mathcal{D}(\mathcal{E}(q_i, \hat{a_i}))$
\State Calculate the loss for the self-reflection:
\State $\mathcal{L}_{reflect}= \sum_{i}^n\textsc{CrossEntropyLoss}(\hat{r_i}, r_i)$
\State Update: $\mathcal{L}_{SR}=\mathcal{L}_{gen}+\mathcal{L}_{reflect} $
\EndWhile
\State Update encoder-decoder parameters $\Phi$
\EndFor
\EndFor
\end{algorithmic}
\end{algorithm}
\subsubsection{Reflective Answering}
Through our preliminary experiments, we found that the standard training process, as described in Eq.~\ref{eqn8}, allows the model to retrieve from documents but also encourages excessive reliance on external knowledge, even when the information is less relevant. This overreliance may result in the model inevitably embedding noise into its parametric knowledge during training. On the other hand, pre-trained MLLMs already possess the required knowledge to directly answer some easier knowledge-intensive questions. Therefore, we propose a novel on-the-fly Reflective-answering mechanism. which trains the generator without external documents while simultaneously generating a self-reflection label. The self-reflection label indicates whether the model considers its answer to be correct based on the context of the question:

\begin{equation}
    p_{\Phi}(\text"correct"|Q, I)=p_{\Phi}(\hat{a} |Q, I)\cdot 
  p_{\Phi}(\text"correct" |Q, I, \hat{a})
\end{equation}

Every time when the answer $\hat{a}$ is generated, the self-reflection label can be generated immediately by comparing $\hat{a}$ with the target answer $\mathcal{A}$. The generated answer will then be concatenated with the query to serve as the context for the binary label prediction -- ``correct'' or ``incorrect''. In this setting, the binary classification task is framed as the next-token prediction, where another casual language modelling loss following the Eq.~\ref{equ8} will be calculated on self-answer and self-reflection generation, denoted as $\mathcal{L}_{SR}$. The joint loss will then be updated with $\mathcal{L}_{SR}$ in Eq.~\ref{eqn11}. Formally, the reflective answering path is described in Alg.~\ref{alg:sr}.
\begin{equation}
      \label{eqn11}
      \mathcal{L}_{Joint}=\mathcal{L}_{R}+\mathcal{L}_{RAG}+\mathcal{L}_{SR}
\end{equation}

\begin{table*}[ht]
\caption{Model performance comparison on OK-VQA. The best performance of our model is highlighted in bold font, and the rows of our models' main results are gray. The best performance in literature is \underline{underlined}. $K$ is the amount of knowledge retrieved in the generation process. Know. Source represents external knowledge source.} 
\begin{center}
\begin{tabular}{l|l l l l | l l}
\hline
    \textbf{No.} & \textbf{Model} & \textbf{Base Models} & \textbf{K} &\textbf{Know. Source} & \textbf{EM(\%)} & \textbf{VQA(\%)}\\
\hline
    \multicolumn{4}{l}{\textit{Classic KB-VQA Systems}} \\
\hline
    1 &KAT-T5&T5-Large&40&Wikipedia&-&44.25\\
    2&TRiG&T5-Large&100&Wikipedia&54.73&50.50\\
    3 &MAVEx&-&-&Wikipeida&-&39.20\\
    4 &RA-VQA&T5-Large&5&GoogleSearch&55.77&51.22\\
    5 &BLIP (zero-shot)&BLIP&-&-&36.99&34.46\\
    6 &BLIP (fine-tuned)&BLIP&-&-&48.89&45.74\\  
    7 &RA-VQA-v2 (FLMR)&BLIP2-$\mathrm{T5_{XL}}$($\sim${3B})&5 &GoogleSearch &\underline{62.01}&60.75\\
\hline
    \multicolumn{4}{l}{\textit{Systems with Large Models (>15B parameters)}} \\
\hline
     8 &PICa&GPT-3 (175B)&-&-&-&48.00\\
    9 &Prophet&GPT-3 (175B)&-&-&-&61.11\\
    10 &REVIVE&GPT-3 (175B)&40&Wikipedia&-&58.00\\
    11 &PALI&PALI (15B)&-&-&-&56.50\\
    12 &Flamingo&Flamingo (80B)&-&-&-&57.80\\
    13 &PaLM-E&PaLM-E (526B)&-&-&-&\underline{66.10}\\
\hline
    \multicolumn{4}{l}{\textit{Base Models without Knowledge Retrieval}} \\
\hline
    14 &InstructBLIP-$\mathrm{T5_{XL}}$ w/o fine-tuned &InstructBLIP-$\mathrm{T5_{XL}}$&-&-& 44.07&41.54\\
    15 &InstructBLIP-$\mathrm{T5_{XL}}$ (fine-tuned)&InstructBLIP-$\mathrm{T5_{XL}}$&-&-& 60.47&55.50\\
    16 &BLIP2-$\mathrm{T5_{XL}}$ w/o fine-tune&BLIP2-$\mathrm{T5_{XL}}$&-&-&12.49&11.60\\
    17 &BLIP2-$\mathrm{T5_{XL}}$ (fine-tuned) &BLIP2-$\mathrm{T5_{XL}}$&-&-& 55.11&52.73\\
    
\hline
    \multicolumn{4}{l}{\textit{Our Proposed Models ($\sim$3B parameters)}} \\
\hline
\multirow{2}{*}{18}&\cellcolor{lightgray!35} UniRVQA (InstructBLIP-$\mathrm{T5_{XL}}$)&\cellcolor{lightgray!35}InstructBLIP-$\mathrm{T5_{XL}}$&\cellcolor{lightgray!35}5&\cellcolor{lightgray!35}GoogleSearch&\cellcolor{lightgray!35}\textbf{66.79}&\cellcolor{lightgray!35}\textbf{61.57}\\
&\multicolumn{4}{l}{\quad \textit{$\%$ relative improvement w.r.t. the base model}}& \textcolor{PineGreen}{\textbf{6.32\%\ $\uparrow$}}
 &\textcolor{PineGreen}{\textbf{6.07\%\ $\uparrow$}}\\
\multirow{2}{*}{19}&\cellcolor{lightgray!35} UniRVQA (BLIP2-$\mathrm{T5_{XL}}$)&\cellcolor{lightgray!35}BLIP2-$\mathrm{T5_{XL}}$&\cellcolor{lightgray!35}5&\cellcolor{lightgray!35}GoogleSearch&\cellcolor{lightgray!35}64.21&\cellcolor{lightgray!35}60.90\\
&\multicolumn{4}{l}{\quad \textit{$\%$ relative improvement w.r.t. the base model}}& \textcolor{PineGreen}{\textbf{9.10\%\ $\uparrow$}}
 &\textcolor{PineGreen}{\textbf{8.17\%\ $\uparrow$}}\\
    
\hline
\end{tabular}
\label{main results}
\end{center}
\end{table*}

\subsection{Inference and Answer Select}
At the inference stage, all documents will first be indexed using PLAID~\cite{Santhanam2022PLAID:Retrieval} for accelerated late-interaction retrieval. The self-reflection mechanism not only enables the model to evaluate its knowledge boundaries but also facilitates adaptive retrieval-augmented generation during the UniRVQA inference stage. At this stage, the model first answers the question without referencing external documents and immediately generate self-reflection prediction to assess the correctness of its response. If the model deems its answer incorrect, the retrieval-augmented process is triggered, allowing it to retrieve documents to facilitate answering. Otherwise, the self-answering result will be kept as the final answer. During the retrieval-augmented generation stage, multiple documents will be selected and generate multiple answers accordingly. Based on Eq.~\ref{eqn1} and ~\ref{eqn2}, the answer with the highest joint probability of retrieval and generation will be selected.

\section{Experiments and Results}
\subsection{Experiments Setup}
\textbf{Datasets.} We mainly evaluate the proposed method on the OK-VQA dataset~\cite{Marino2019OK-VQA:Knowledge} and conduct complementary experiments using the InfoSeek dataset~\cite{Chen2023CanQuestions}. InfoSeek is regarded as a more knowledge-dependent dataset, as its questions are often unanswerable without external knowledge support. Here are the details about two datasets: 

(1) OK-VQA~\cite{Marino2019OK-VQA:Knowledge} dataset contains over 10k questions on MSCOCO \cite{Lin2014MicrosoftContext} images which require external knowledge to answer. For the external knowledge base we adopt the Google Search Corpus~\cite{Luo2021Weakly-SupervisedAnswering}, which is a textual corpus containing 166,389 passages from Google, covering all the knowledge necessary for answering questions in OK-VQA. We use the original splits of training and testing sets to ensure comparability, and use 10\% of the training set for validation. 

(2) InfoSeek~\cite{Chen2023CanQuestions} is a newly proposed large-scale KB-VQA dataset introduced in 2023, built on the OVEN image dataset~\cite{Hu2023Open-domainEntities}. Following the original paper we use Wikipedia~\cite{Vrandecic2014Wikidata} as the external knowledge base. Compared to OK-VQA, InfoSeek is more challenging as it encompasses a greater number of questions that necessitate expertise knowledge for accurate responses. Given the large size (over 1 million image-question pairs) of InfoSeek and Wikipedia corpus, we conduct our experiments on a down-sampled subset following existing works~\cite{Lin2024PreFLMR:Retrievers}. See supplementary material for more details.

\noindent\textbf{Implementation Details.} We select BLIP2-Flan-T5-XL~\cite{Li2023BLIP-2:Models} and InstructBLIP-Flan-T5-XL~\cite{Dai2023InstructBLIP:Tuning} as the MLLM base models to build our proposed framework. We use 1 Nvidia A100 with 80GB VRM for all experiments. We use DoRA~\cite{Liu2024DoRA:Adaptation} to fine-tune UniRVQA. We choose a batch size of 20 and the AdamW optimizer~\cite{Loshchilov2017DecoupledRegularization} with the learning rate set as 2e10-4. The scheduler modulates the learning rate throughout the training process, starting with a warmup period of 100 steps before gradually reducing the learning rate following a cosine schedule. To ensure comparability and avoid randomness bias, we report our main results as the average from 3 different random seed settings.

\noindent\textbf{Evaluation.} We present the metrics used to assess answer generation and knowledge retrieval performance:

(1) \textit{Exact Match (EM):} We evaluate the exact matching between the generated answer and the answer set $S$, where $\#s(\hat{a})$ is the occurrences of $\hat{a}$ in $S$: 
\begin{equation}
    \mathrm{EM}(\hat{a}, S)=\mathrm{min}(\#s(\hat{a}),1)
\end{equation}

(2) \textit{VQAScore:} on OK-VQA dataset, we use an additional official VQA Score~\cite{Marino2019OK-VQA:Knowledge}. This score makes the model partially rewarded if it generates a less popular answer among human responses: 
\begin{equation}
    \mathrm{VQAScore}(\hat{a}, S)=\mathrm{min}(\#s(\hat{a})/3,1)
\end{equation}

(3) \textit{Pseudo Relevance Recall (PRR@K):} Following previous work ~\cite{Lin2022RetrievalKnowledge, Lin2023Fine-grainedAnswering}, We adopt pseudo-relevance labels and evaluate retrieval performance by counting the number of questions whose top-K retrieved documents contain the correct answers. Following prior work, we set K=5 in the main experiments.

\noindent\textbf{Baseline Models.} We compare our proposed framework with the latest KB-VQA systems in answer generation performance. Among them, the first group of systems smaller model with less than 3B parameters, including: KAT~\cite{Gui2021KAT:Vision-and-Language}, TRiG~\cite{Gao2022Transform-Retrieve-Generate:Answering}, MAVEx~\cite{Wu2022Multi-ModalVQA}, RA-VQA~\cite{Lin2022RetrievalKnowledge}, BLIP~\cite{Li2022BLIP:Generation}, and FLMR~\cite{Lin2023Fine-grainedAnswering}. The second group includes larger systems that are built with large pre-trained models such as GPT-3 (175B)~\cite{Brown2020LanguageLearners} and PaLM-E (526B):
PICa~\cite{Yang2022AnVQA}, Prophet~\cite{Yu2023Prophet:Answering}, REVIVE~\cite{Lin2022REVIVE:Answering}, PALI~\cite{Chen2023PaLI:Model}, Flamingo~\cite{Alayrac2022Flamingo:Learning} and PaLM-E~\cite{Driess2023PaLM-E:Model}.
Additionally, we will compare our retrieval performance against strong retrieval-focused models designed specifically for VQA task, mainly including Dense Passage Retrieval (DPR)~\cite{Karpukhin2020DenseAnswering}, FLMR~\cite{Lin2023Fine-grainedAnswering} and preFLMR~\cite{Lin2024PreFLMR:Retrievers}.

\begin{table}[t]
\caption{VQA performance comparison on original InfoSeek validation split with K=5. Un-Q and Un-E stand for two types of test questions -- unseen questions and unseen entities.}
\begin{center}
\begin{tabular}{l| l l l l}
\hline
    && \multicolumn{3}{l}{\textbf{Accuracy (\%)}}\\\cmidrule{3-5}    
     \textbf{Model}  &Base Models & Un-Q & Un-E & All \\
\hline
\multicolumn{4}{l}{\textit{Standard fine-tuned on the dataset}} \\
\hline
PaLM(Q-only)&PaLM&5.5&4.2&4.8\\
BLIP2-$\mathrm{T5_{XL}}$&BLIP2&12.7&12.3&12.5\\
InstructBLIP-$\mathrm{T5_{XL}}$&InstructBLIP&15.0&14.0&14.5\\
PALI-17B&PALI&24.2&16.7&19.7\\
\hline
\multicolumn{4}{l}{\textit{Fine-tuned with knowledge}}\\
\hline
CLIP + PaLM&PaLM(540B)&22.7&18.5&20.4\\
CLIP + FiD&-&23.3&19.1&20.9\\
UniRVQA&InstructBLIP&\textbf{24.01}&\textbf{20.40}&\textbf{22.06}\\
\hline
\multicolumn{2}{l}{\textit{$\%$ improv. w.r.t. the base model}}&\textcolor{PineGreen}{\textbf{9.01\%\ $\uparrow$}}&\textcolor{PineGreen}{\textbf{6.40\%\ $\uparrow$}}&\textcolor{PineGreen}{\textbf{7.56\%\ $\uparrow$}}\\
  
\hline
\end{tabular}
\label{infoseek2}
\end{center}
\end{table}
\begin{figure}
    \centering
    \includegraphics[width=\columnwidth]{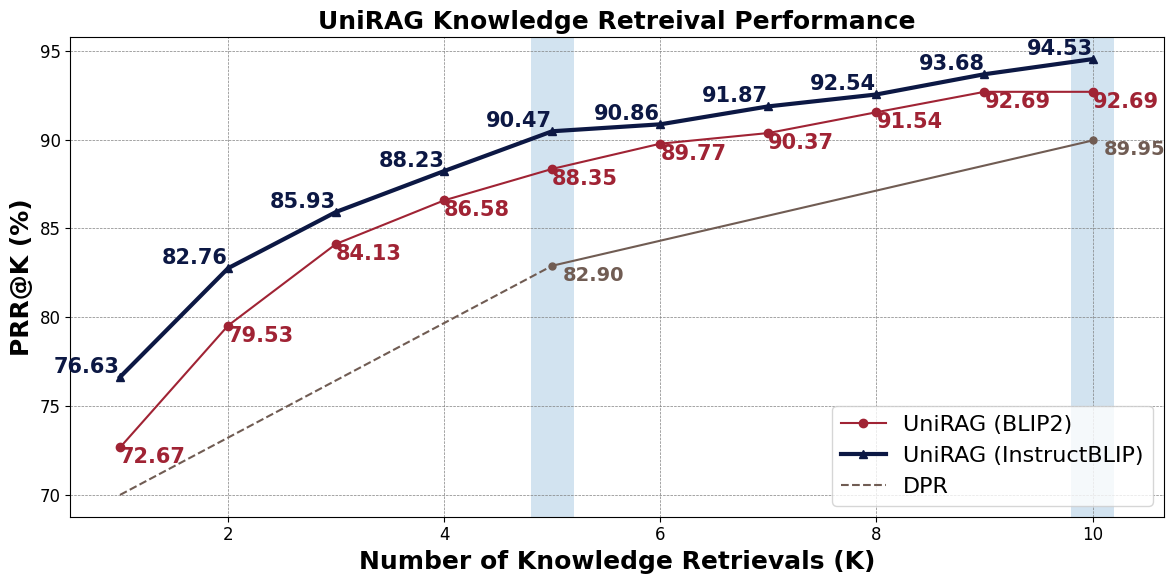}
    \caption{Retrieval performance variation with respect to the number of retrieved knowledge evaluated on OK-VQA.The DPR results shown are for baseline reference.}
    \label{fig:number of retreived knowlwedge}
\end{figure}

\subsection{Main Results and Analysis}
According to the experiment results, our key observations are:
 \begin{itemize}
    \item Our proposed model achieves the state-of-art performance in both answer generation and knowledge retrieval. It achieves the highest EM of 66.79\% on the OK-VQA dataset, with a notable 4.78\% improvement over the best model in literature. It also delivers the best knowledge retrieval results across datasets, with an average improvement over 3\%. 
    \item UniRVQA significantly and consistently improves the VQA performance of base MLLMs, achieving an average accuracy gain of 7.42\% and 7.66\% on two datasets, demonstrating its ability to extend the effectiveness of existing MLLMs in knowledge-intensive tasks.
    \item Compared to other high-performing very large models, such as PaLM-E (526B)\cite{Driess2023PaLM-E:Model} and PaLI (15B) \cite{Chen2023PaLI:Model}, UniRVQA achieves competitive performance with a compact size (3B) and a training time of under 3 hours, demonstrating its high efficiency without sacrificing performance.  
\end{itemize}

\subsubsection{VQA Performance}
The overall accuracy performance comparison between our models and the baseline models on the OK-VQA dataset is shown in Table~\ref{main results}. First, our proposed framework implemented with various base models (InstructBLIP~\cite{Dai2023InstructBLIP:Tuning} and BLIP2~\cite{Li2023BLIP-2:Models}) achieves the top-tier performance. Specifically, UniRVQA (InstructBLIP) delivers the best EM accuracy, with a 4.78\% improvement over the previous best model (FLMR~\cite{Lin2023Fine-grainedAnswering}).  Additionally, it achieves a highly competitive VQAScore accuracy of 61.57\%. Meanwhile, UniRVQA (BLIP2) secures second place in EM (64.21\%) and maintains strong VQA accuracy at 60.90\%. UniRVQA also demonstrates exceptional performance in complementary experiments on InfoSeek (Table~\ref{infoseek2}). It achieves the highest answering accuracy of 24.01\% on Unseen Questions and 20.40\% on Unseen Entities.

\textbf{Boosting base MLLMs performance.} While extremely large models with more than 15B parameters(Table~\ref{main results}, lines 9–13) are initially advantageous in VQA performance and benefit from their capacity to store extensive knowledge through large-scale training, UniRVQA effectively closes the performance gap between them and smaller base models (3B) (Table~\ref{main results}, lines 14–17). By integrating UniRVQA, the base MLLMs achieve leading performance levels. Compared to their standard fine-tuned counterparts, the variants of UniRVQA deliver an average improvement of 7.42\% (Table~\ref{main results}, lines 18–19 vs. lines 15, 17). The similar observation can be concluded from the InfoSeek dataset (Table~\ref{infoseek2}), where UniRVQA raises the base model's overall accuracy from 14.5\% to 22.1\%, demonstrating an impressive 7.56\% gain. The outcomes highlight UniRVQA can greatly unlock the potential of general MLLMs in addressing knowledge-intensive VQA tasks. Its adaptable framework not only maximizes the utility of existing models but also paves the way for leveraging future advancements in MLLMs. Additionally, UniRVQA proves especially advantageous in scenarios with limited access to extremely large models or constrained computational resources.

Lastly, we would like to point out the performance on CLIP combined with FiD (Table~\ref{infoseek2}), as referred from the original paper~\cite{Chen2023CanQuestions}, achieves the second-best result using a special setting of retrieving 100 documents per question, which is far exceeds 5 documents used by ours and most of baseline methods. Obtaining better performance under less favorable settings  again highlights the effectiveness of our approach. Furthermore, our model achieves competitive results with remarkable efficiency, requiring only 3 GPU hours for 3,000 training steps, in contrast to other models that typically demand over 24 GPU hours~\cite{Lin2024PreFLMR:Retrievers, Lin2023Fine-grainedAnswering, Driess2023PaLM-E:Model, Chen2023PaLI:Model}.

\subsubsection{Retrieval Performance}
Besides question-answering performance, we evaluate the model's retrieval performance to understand the system's ability in leveraging external resources. As shown in Table~\ref{infoseek}, UniRVQA (InstructBLIP) achieves state-of-the-art retrieval performance, with PRR@5 scores of 90.47\% on OK-VQA and 68.51\% on InfoSeek. While UniRVQA maintains its leading position on OK-VQA, we observe relatively small performance gaps between models on this simpler dataset, with UniRVQA surpassing the previous best model by only around 1\% across retrieval levels. However, on the more challenging InfoSeek dataset, our model demonstrates significantly superior robustness, achieving an 8.91\% improvement over existing methods, which typically struggle with the performance drops. This underscores UniRVQA's strong capability to tackle complex, knowledge-intensive scenarios.

Additionally, we analyze how retrieval performance evolves with the number of retrieved knowledge passages. Figure~\ref{infoseek} shows that UniRVQA (InstructBLIP and BLIP2) achieves high recall earlier in the retrieval process, reaching approximately 80\% recall with only the top-2 passages, whereas the baseline requires the top-5 passages for comparable performance. Beyond this, UniRVQA improves sharply by 14\% as K increases from 1 to 5, after which improvements taper off. This demonstrates that the model efficiently retrieves relevant documents, reducing the computational burden of excessive retrievals. Furthermore, UniRVQA's superior retrieval performance highlights that, with the unified framework, reasoning skills acquired during answer generation can positively impact retrieval. This synergy is further explored in the ablation study.

\begin{table}
\caption{Retrieval performance comparison on InfoSeek and OK-VQA. We report only PRR@5 on InfoSeek to follow the previous work. The best performance of our model is highlighted in bold font.}
\begin{center}
\begin{tabular}{l| c c c c}
\hline
\hline
    &\textbf{InfoSeek}& \multicolumn{2}{c}{\textbf{OK-VQA}}\\  
     \textbf{Model}  &PRR@5 & PRR@5 & PRR@10 \\
\hline
DPR&44.88&82.90&89.95\\
FLMR&46.42&89.32&94.00\\
PreFLMR&59.60&-&-\\
\hline
UniRVQA(BLIP2)&63.65&88.35&92.69\\
UniRVQA(InstructBLIP)&\textbf{68.51}&\textbf{90.47}&\textbf{94.53}\\
\hline
\hline
\end{tabular}
\label{infoseek}
\end{center}
\end{table}

\subsection{Ablation Study}
We analyze the effectiveness of the main components in our proposed framework to answer the following research questions. The ablation study is conducted on the OK-VQA dataset.
\begin{figure}
    \centering
    \includegraphics[width=\columnwidth]{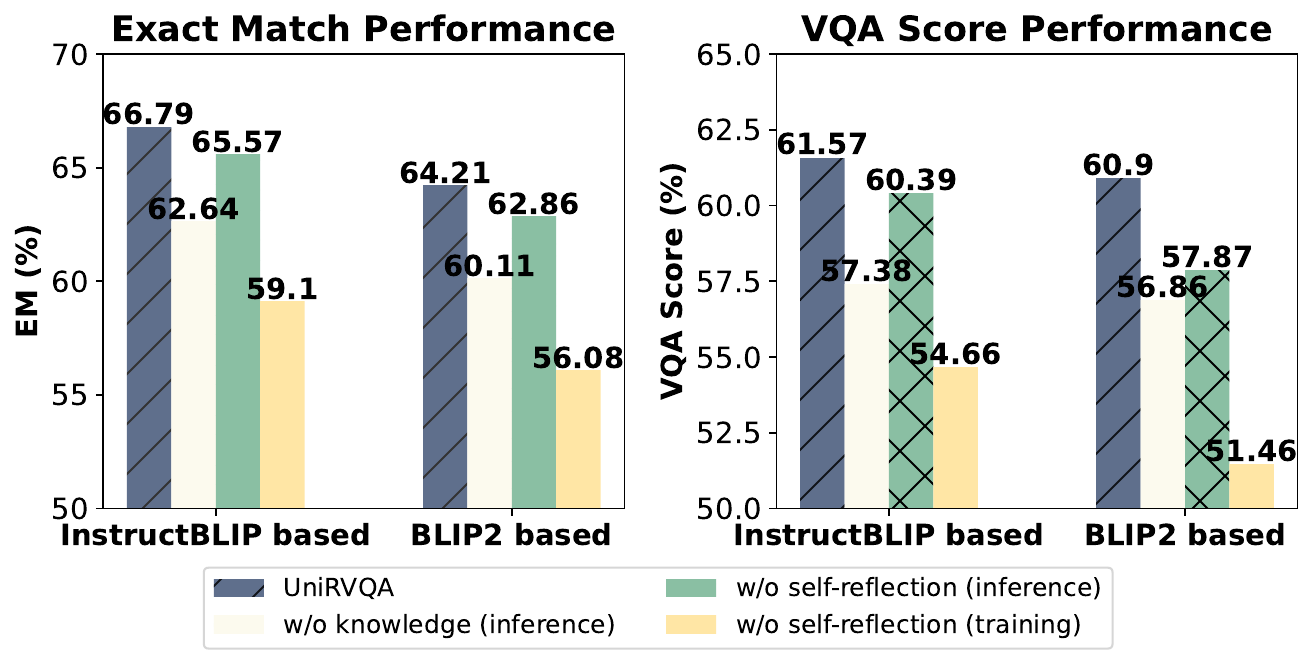}
    \caption{Ablation study on the self-reflection mechanism. Two groups of models in each graph are based on InstructBLIP and BLIP2 respectively.}
    \label{ablation study fig}
\end{figure}

\textbf{RQ1: How well does collaborative parametric knowledge calibration perform}? To answer the question, we construct two variants of our model by separating the training process of retriever and answer generator in different ways. We keep the late interaction and reflective-answering mechanism in all variants. The results are summarized in Table~\ref{tab2}. Specifically, $\mathrm{UniRVQA^{--}}$ employs the traditional setting of separating the retriever and the generator. By training the models separately, $\mathrm{UniRVQA^{--}}$ performs worse than our proposed $\mathrm{UniRVQA}$. Compared to the setting where two MLLMs handle knowledge-intensive tasks separately, our proposed unified framework enables retriever and answer generator to share the same network, reducing the model size significantly and allowing both tasks to complement each other, resulting in improved performance.

Additionally, $\mathrm{UniRVQA^{-}}$ uses a unified framework but without collaborative knowledge calibration, where the network is first trained on the retrieval task, followed by the answer generation task. The results show that simply unifying the framework without proper training strategy design leads to a significant drop in performance. This suggests that naive multistage training may hinder the model’s ability to share parametric knowledge effectively. The better performance on UniRVQA verifies our assumption that by performing collaborative parametric knowledge calibration, the model can better leverage capabilities learned from one task to improve the performance on the other.


\begin{figure*}[]
    \centering  \includegraphics[width=\textwidth]{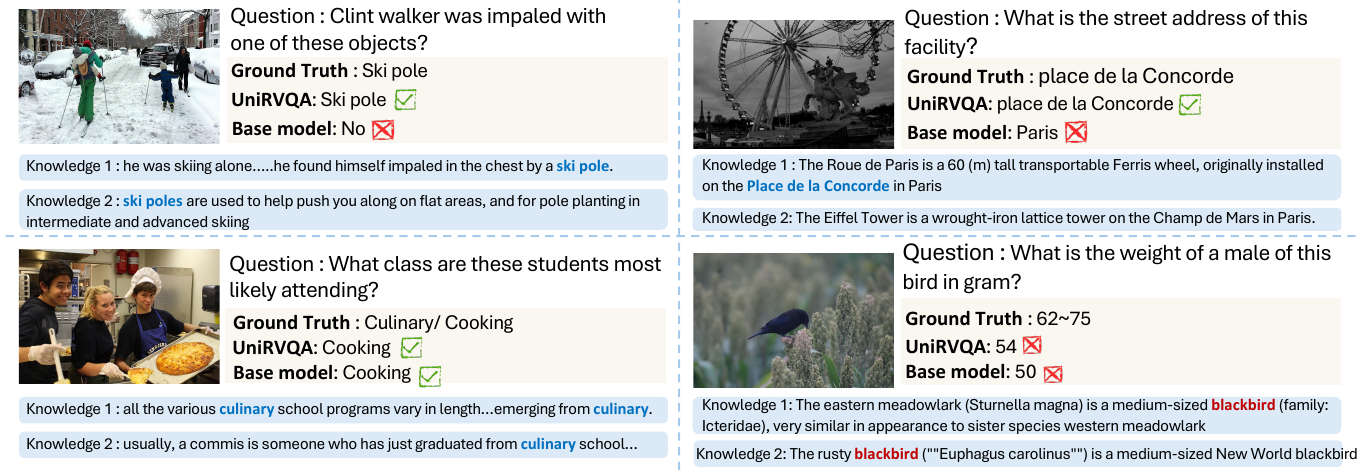}
    \caption{Qualitative results on four cases. The left two cases are from OK-VQA and the right two cases are from InfoSeek. UniRVQA refers to UniRVQA (InstructBLIP-$\mathrm{T5_{XL}}$) and baseline refers to the origin InstructBLIP-$\mathrm{T5_{XL}}$. For the limit of space, we only present the top-2 retrieve results here.}
    \label{qualitative}
\end{figure*}

\begin{table}[t]
\centering
\caption{Model variation performances with K=5. Base model is built by InstructBLIP-$\mathrm{T5_{XL}}$. K.S. and C.C. mean knowledge sharing and collaborative calibration.}
\renewcommand{\arraystretch}{1.1}
\setlength{\tabcolsep}{4pt}
\begin{tabularx}{\linewidth}{lccccc}
\toprule
\textbf{Model} & \textbf{K. S.} & \textbf{C. C.} & \textbf{PRR@5} & \textbf{VQA (\%)} & \textbf{EM (\%)} \\
\midrule
$\mathrm{UniRVQA^{--}}$  & -- & -- & 86.45 & 57.97 & 60.01 \\
$\mathrm{UniRVQA^{-}}$   & \checkmark & -- & 76.00 & 49.13 & 51.38 \\
\rowcolor{gray!20}
UniRVQA                 & \checkmark & \checkmark & 90.47 & 61.57 & 66.79 \\
\bottomrule
\end{tabularx}
\label{tab2}
\end{table}
\textbf{RQ2: How can the Reflective-Answering enhance the system performance?}
To investigate the impact of reflective-answering mechanism, we conduct experiments by disabling the mechanism during the training and inference stage respectively. The results presented in Fig.~\ref{ablation study fig}, show that removing reflective-answering leads to a significant drop in performance compared to the complete UniRVQA model. Specifically, removing self-reflection during training results in the most substantial decrease, with a average reduction of over 6.91\% in VQA score and more than 7.5\% in EM for both base models. This highlights the critical role of self-reflection in improving model performance during training. Additionally, turning off self-reflection during inference hinders performance by 1.3\% on average, demonstrating that reflective-answering is crucial not only for learning but also for making more accurate predictions. Specifically, incorporating the reflective-answering mechanism into inference yields accuracy gains ranging from 1.22\% to 3.03\%. We attribute the effectiveness of our proposed mechanism to its ability to alleviate the model's dependence on external knowledge. With reflective-answering, the model is more discerning in its use of external information, avoiding reliance on irrelevant data when it already possesses implicit knowledge to answer accurately. 

\textbf{RQ3: How does the fine-grained knowledge learning (late-interaction) support the system?} Here, we further investigate the contributions of fine-grained knowledge in our experiments by removing them during the inference. As shown in Figure~\ref{ablation study fig}, removing the external knowledge support will directly lead to an average reduction of 4.1\% in answer accuracy, which confirms the necessity of fine-grained external knowledge processing when answering knowledge-intensive questions. Compared to the standard fine-tuning method where only the final answer is provided as training labels (Table~\ref{main results}, line 15 and 17), the UniRVQA framework improves answer accuracy by 2\% on average. Specifically, with the retrieval augmentation turned off, UniRVQA (InstructBLIP) improves the fine-tuned counterparts from 60.47\% to 62.64\% in EM and from 55.50\% to 57.38\% in VQAScore. Such improvement indicates that incorporating the fine-grained knowledge learning during the training process appropriately can be beneficial to the reasoning ability of the answer generator. 

\subsection{Case Studies}
We conduct a qualitative study using InstructBLIP-$\mathrm{T5_{XL}}$ as the base model, with results visualized in Fig.~\ref{qualitative}. The left column shows two successful cases from OK-VQA. In the first example, UniRVQA retrieves documents describing ``ski poles'', precisely addressing a question outside the base model's implicit knowledge. The other example demonstrates a case where the model confidently identified that it could answer the question without the need for retrieval, thus saving inference time. We also display the retrieval results here, which are highly relevant and provide supporting information about culinary school. Even if the retrieved documents were irrelevant, the reflective-answering mechanism could ensure that our model would not be affected by the noises from those documents.

The right column features two examples from InfoSeek. The top-right example asks for the specific location where the facility is, a question that is hard to answer without particular external knowledge. This also highlights the complexity of questions in InfoSeek. UniRVQA accurately retrieves an encyclopedic document describing the ferris wheel installed on the ``Place de la Concorde'' in Paris, showcasing the model's ability to effectively identify and use fine-grained information to answer challenging questions. We also note that although the base model was not able to provide the precise location, it could still identify the city in the image, which indicates that the base MLLM contains some fundamental knowledge that can be potentially leveraged. The bottom-right failure case is challenging. The model struggled with identifying the species of the black bird in the image, however still managed to retrieve generally relevant information about ``blackbird''. We emphasize that a potential area for improvement is the fine-grained entity retrieval, particularly in distinguishing visually similar entities.

\section{Conclusion}
In this work, we present a unified retrieval-augmented KB-VQA framework with collaborative parametric knowledge calibration to address the limitations of decoupled retrieval and generation stages. By enabling parametric knowledge sharing between retriever and generator, and integrating both a late-interaction mechanism for fine-grained multimodal understanding and a reflective-answering mechanism for knowledge boundary assessment, our training framework significantly improves the adaptability of general MLLMs to knowledge-intensive tasks. Extensive experiments demonstrate that our approach not only achieves state-of-the-art performance but also substantially boosts the capabilities of base MLLMs, which offers a practical and efficient solution for future KB-VQA research.


\bibliographystyle{ACM-Reference-Format}
\bibliography{references}

\end{document}